\documentclass{svproc}
\usepackage{url}
\usepackage{amsfonts}

\usepackage{amsmath}
\usepackage{amssymb,amsbsy}
\usepackage{amsthm}
\usepackage{graphicx}
\usepackage{xcolor}
\usepackage{comment}
\usepackage[normalem]{ulem}
\usepackage[utf8]{inputenc}
\usepackage[english]{babel}

\DeclareMathOperator*{\argmax}{arg\,max}

\newcommand{\zmk}[1]{{\color{magenta} #1}}

\begin{document}
\mainmatter              
\title{Using Reinforcement Learning to Herd a Robotic Swarm to a Target Distribution}
\titlerunning{DARS 2021}  
%
\author{Zahi Kakish\inst{1} \and Karthik Elamvazhuthi\inst{2} \and Spring Berman\inst{1}}
\authorrunning{Zahi Kakish et al.} 
%
\tocauthor{Ivar Ekeland, Roger Temam, Jeffrey Dean, David Grove,
Craig Chambers, Kim B. Bruce, and Elisa Bertino}
\institute{Arizona State University, Tempe AZ 85281, USA,\\
\email{zahi.kakish@asu.edu, spring.berman@asu.edu},\\ 
\and
University of California, Los Angeles CA 90095, USA,\\
\email{karthikevaz@math.ucla.edu}
}

\maketitle              

\begin{abstract}
In this paper, we present a reinforcement learning approach to designing a control policy for a ``leader'' agent that herds a swarm of ``follower'' agents, via repulsive interactions, as quickly as possible to a target probability distribution over a strongly connected graph. The leader control policy is a function of the swarm distribution, which evolves over time according to a mean-field model in the form of an ordinary difference equation. The dependence of the policy on agent populations at each graph vertex, rather than on individual agent activity, simplifies the observations required by the leader and enables the control strategy to scale with the number of agents. Two Temporal-Difference learning algorithms, SARSA and Q-Learning, are used to generate the leader control policy based on the follower agent distribution and the leader's location on the graph. A simulation environment corresponding to a grid graph with 4 vertices was used to train and validate the control policies for follower agent populations ranging from 10 to 1000. 
Finally, the control policies trained on 100 simulated agents were used to successfully redistribute a physical swarm of 10 small robots to a target  distribution among 4 spatial regions. 

\end{abstract}

\section{Introduction}


We present two Temporal-Difference learning algorithms \cite{sutton2018reinforcement} 
for generating a control policy that guides a mobile agent, referred to as a {\it leader}, to herd a swarm of autonomous {\it follower} agents to a target distribution among a small set of states. This leader-follower control approach can be used to redistribute a swarm of low-cost robots with limited capabilities and information using a single robot with sophisticated sensing, localization, computation, and planning capabilities, in scenarios where the leader lacks a model of the swarm dynamics. Such a control strategy is useful for many applications in swarm robotics, including exploration, environmental monitoring, inspection tasks, disaster response, and targeted drug delivery at the micro-nanoscale.

There has been a considerable amount of work on leader-follower multi-agent control schemes in which the leader has an attractive effect on the followers \cite{ji2008containment,mesbahi2010graph}. Several recent works have presented models for herding robotic swarms using leaders that have a repulsive effect on the swarm \cite{pierson2017controlling,elamvazhuthi2016confinement,paranjape2018robotic}. Using such models, analytical controllers for herding a swarm have been constructed for the case when there is a single leader \cite{elamvazhuthi2016confinement,paranjape2018robotic} and multiple leaders \cite{pierson2017controlling}. The controllers designed in these works are not necessarily optimal for a given performance metric. To design optimal control policies for a herding model, the authors in \cite{go2016reinforcement} consider a reinforcement learning (RL) approach. While existing herding models are suitable for the objective of confining a swarm to a small region in space, many applications require a swarm to cover an area according to some target probability density. If the robots do not have spatial localization capabilities, then the controllers developed in \cite{ji2008containment,mesbahi2010graph,pierson2017controlling,elamvazhuthi2016confinement,paranjape2018robotic,go2016reinforcement} cannot be applied for such coverage problems. Moreover, these models are not suitable for herding large swarms using RL-based control approaches, since such approaches would not scale well with the number of robots. This loss of scalability is due to the fact that the models describe individual agents, which may not be necessary since robot identities are not important for many swarm applications.
 
In this paper, we consider a {\it mean-field} or {\it macroscopic} model that describes the swarm of follower agents as a probability distribution over a graph, which represents the configuration space of each agent. Previous work has utilized similar mean-field models to design a set of control policies that is implemented on each robot in a swarm in order to drive the entire swarm to a target distribution, e.g. 
for problems in 
spatial coverage and task allocation
\cite{elamvazhuthi2019mean}. In this prior work, all the robots must be reprogrammed with a new set of control policies if the target distribution is changed. In contrast, our approach can achieve new target swarm distributions via redesign of the control policy of a single leader agent, while the control policies of the swarm agents remain fixed. 
The 
follower agents switch stochastically out of their current location on the graph whenever the leader is at their location; in this way, the leader has a ``repulsive'' effect on the followers. The transition rates out of each location are common to all the followers, and are therefore independent of the agents' identities. Using the mean-field model, herding objectives for the swarm are framed in terms of the distribution of the followers over the graph. The objective is to compute leader control policies that are functions of the agent distribution, rather than the individual agents' states, which makes the control policies scalable with the number of agents. \zmk{
}

We apply RL-based approaches to the mean-field model to construct leader control policies that minimize the time required for the swarm of follower agents to converge to a user-defined target distribution. The RL-based control policies are not hindered by curse-of-dimensionality issues that arise in classical optimal control approaches. Additionally, RL-based approaches can more easily accommodate the stochastic nature of the follower agent transitions on the graph. There is prior work on RL-based control approaches for mean-field models of swarms in which each agent can localize itself in space and a  state-dependent control policy can be assigned to each agent directly \cite{vsovsic2018reinforcement,huttenrauch2019deep,yang2018mean}. However, to our knowledge, there is no existing work on RL-based approaches applied to mean-field models for herding a swarm using a leader agent.
Our approach provides an RL-based framework for designing scalable strategies to control swarms of  resource-constrained robots using a single leader robot, and it can be extended to other types of swarm control objectives. 


\section{Methodology}
\label{sec:methodology}

\subsection{Problem Statement}
\label{sec:mean_field_model}

We first define some notation from graph theory and matrix analysis that we use to formally state our problem. We denote by $\mathcal{G} = (\mathcal{V},\mathcal{E})$ a directed graph with a set of $M$ vertices, $\mathcal{V} = \lbrace 1,...,M \rbrace$, and a set of $N_{\mathcal{E}}$ edges, $\mathcal{E} \subset \mathcal{V} \times \mathcal{V}$, where $e = (i,j) \in \mathcal{E}$ if there is an edge from vertex $i \in \mathcal{V}$ to vertex $j \in \mathcal{V}$. We define a {\it source map} $\sigma : \mathcal{E} \rightarrow \mathcal{V}$  and a {\it target map} $\tau: \mathcal{E} \rightarrow \mathcal{V}$ for which $\sigma(e) = i$ and $\tau(e) = j$ whenever $e = (i,j) \in \mathcal{E}$. Given a vector $X  \in \mathbb{R}^M$, $X_i$ refers to the $i^{th}$ coordinate value of $X$. For a matrix $A \in \mathbb{R}^{M \times N}$, $A^{ij} $  refers to the element in the $i^{th}$ row and $j^{th}$ column of $A$.

We consider a finite swarm of $N$ follower agents and a single leader agent. The locations of the leader and followers evolve on a graph, $\mathcal{G} = (\mathcal{V}, \mathcal{E})$, where $\mathcal{V} = \lbrace 1,...,M \rbrace$ is a finite set of vertices and $\mathcal{E} = \{(i,j)~|~ i, j \in \mathcal{V}\}$ is a set of edges that define the pairs of vertices between which agents can transition. The vertices in $\mathcal{V}$ represent a set of spatial locations obtained by partitioning the agents' environment. We will assume that the graph $\mathcal{G} = (\mathcal{V}, \mathcal{E})$ is strongly connected and that there is a self-edge $(i,i)\in \mathcal{E}$ at every vertex $i \in \mathcal{V}$. We assume that the leader agent can count the number of follower agents at each vertex in the graph. The follower agents at a location $v$  only decide to move to an adjacent location if the leader agent is currently at location $v$ and is in a particular behavioral state. In other words, the presence of the leader \textit{repels} the followers at the leader's location. The leader agent does not have a model of the follower agents' behavior.

The leader agent performs a sequence of transitions from one location (vertex) to another. The leader's location at time $k \in \mathbb{Z}_+$ is denoted by $\ell_1(k) \in \mathcal{V}$. In addition to the spatial state $\ell_1(k)$, the leader has a behavioral state at each time $k$, defined as $\ell_2(k) \in \lbrace 0,1 \rbrace$. 
The location of each follower agent $i \in \{1,...,N\}$ is defined by a discrete-time Markov chain (DTMC) $X_i(k)$ that evolves on the state space $ \mathcal{V}$ according to the conditional probabilities
\begin{equation} \label{eq:DTMC}
	\mathbb{P}(X_{i}(k + 1) = \tau(e)~|~X_{i}(k) = \sigma(e)) = u_{e}(k)
\end{equation}

For each $v\in \mathcal{V}$ and each $e\in \mathcal{E}$ such that $\sigma(e)=v \neq \tau(e)$, $u_e(k)$ is given by 
\begin{equation}
	u_{e}(k) =
	\begin{cases}
		& ~ \beta_e  \quad  \text{if  }~   \ell_1(k)=\sigma(e) ~\text{ and }~ \ell_2(k)=  1,  \\
		& ~0  \quad~~ \text{if  }~ \ell_1(k)  = \sigma(e)  ~\text{ and }~ \ell_2(k)=  0,\\
		& ~ 0  \quad~~ \text{if  }~ \ell_1(k)  \neq \sigma(e), 
	\end{cases}
\end{equation}
where $\beta_e$ are positive parameters such that  $\sum_{\substack{e \in \mathcal{E} \\ v=\sigma(e) \neq \tau(e)}} \beta_e <1$. Additionally, for each $v\in \mathcal{V}$, $u_{(v,v)}(k)$ is given by 
\begin{equation}
    u_{(v,v)}(k) = 1 ~ - \sum_{\substack{e \in \mathcal{E} \\ v=\sigma(e) \neq \tau(e)}} u_e(k) 
\end{equation}

For each vertex $v \in \mathcal{V}$, we define a set of possible actions $A_v$ taken by the leader when it is located at $v$:
\begin{equation}
	\label{eq:action_set}
	A_v = \bigcup_{\substack{e \in \mathcal{E} \\ v=\sigma(e)}}\lbrace e \rbrace \times \lbrace 0,1 \rbrace
\end{equation}
The leader transitions between states in $\mathcal{V} \times \lbrace 0,1 \rbrace$ according to the conditional probabilities
\begin{equation}
\label{eq:prob_agent_controller}
\mathbb{P}(\ell_1(k + 1) = \tau(e), \ell_2(k + 1)= d ~|~ \ell_1(k) = \sigma(e))= 1
\end{equation}
if $p(k)$, the action taken by the leader at time $k$ when it is at vertex $v$, is given by $p(k) = (e,d) \in A_v $.

The fraction, or {\it empirical distribution}, of follower agents that are at location $v\in \mathcal{V}$ at time $k$ is given by  $\frac{1}{N}\sum_{i=1}^N\chi_v(X_i(k))$, where $\chi_v(w) = 1$ if $w = v$ and $0$ otherwise. 
Our goal is to learn a policy that navigates the leader between vertices using the actions $p(k)$ such that the follower agents are redistributed (``herded'') from their initial empirical distribution $\frac{1}{N}\sum_{i=1}^N\chi_v(X_i(0))$ among the vertices to a desired empirical distribution $\frac{1}{N}\sum_{i=1}^N\chi_v(X_i(T))$ at some final time $T$, where $T$ is as small as possible. Since the identities of the follower agents are not important, 
we aim to construct a control policy for the leader that is a function of the current empirical distribution $\frac{1}{N}\sum_{i=1}^N\chi_v(X_i(k))$, rather than the individual agent states $X_i(k)$. However, $\frac{1}{N}\sum_{i=1}^N\chi_v(X_i(k))$ is not a state variable of the DTMC. In order to treat $\frac{1}{N}\sum_{i=1}^N\chi_v(X_i(k))$ as the state, we consider the {\it mean-field limit} of this quantity as $N \rightarrow \infty$. Let  $\mathcal{P}(\mathcal{V})= \lbrace Y \in \mathbb{R}^M_{\geq 0}; ~\sum_{v=1}^M Y_v = 1 \rbrace$ be the simplex of probability densities on $\mathcal{V}$. When $N \rightarrow \infty$, the empirical distribution $\frac{1}{N}\sum_{i=1}^N\chi_v(X_i(k))$ converges to a deterministic quantity $\hat{S}(k) \in \mathcal{P(V)}$, which evolves according to the following {\it mean-field model}, a system of difference equations that define the discrete-time  
Kolmogorov Forward Equation:
\begin{equation}
	\label{eq:ctrsys}
	\hat{S}(k+1) = \sum_{ e\in \mathcal E} u_e(k) B_e \hat{S}(k), 
	~~~\hat{S}(0) = \hat{S}^0 \in \mathcal{P}(\mathcal{V}),
\end{equation}
where $B_e$ are matrices whose entries are given by 
\[
B_e^{ij} = 
\begin{cases} 
1 & \text{if } i= \tau(e), \hspace{1mm} j = \sigma(e),\\
0       & \text{otherwise.}
\end{cases}
\] 
The random variable $X_i(k)$ is related to the solution of the difference equation \eqref{eq:ctrsys} by the relation $\mathbb{P}(X_i(k) = v) = \hat{S}_v(k)$. 

We formulate an optimization problem that minimizes the number of time steps $k$ required for the follower agents to converge to $\hat{S}_{target}$, the target distribution. In this optimization problem, the reward function is defined as
\begin{equation}
    \label{eq:reward}
    R(k) = -1 \cdot \mathbb{E}||\hat{S}(k) - \hat{S}_{target} ||^2.
\end{equation}

\begin{problem}
Given a target follower agent distribution $\hat{S}_{target}$, devise a leader control policy $\pi: \mathcal{P(V)} \times \mathcal{V} \rightarrow A$ that drives the follower agent distribution to $\hat{S}(T) = \hat{S}_{target}$, where the final time $T$ is as small as possible, by minimizing the total reward $\sum_{k=1}^T R(k)$. The leader action at time $k$ when it is at vertex $v \in \mathcal{V}$ is defined as $p(k)=\pi(\hat{S}(k),\ell_1(k)) \in A_v$ for all $k \in \lbrace 1,...,T \rbrace$, where 
$A = \cup_{v \in \mathcal{V}}A_v$.
\end{problem} 

\subsection{Design of Leader Control Policies using Temporal-Difference Methods}
\label{sec:temporal_difference_controller}

Two Temporal-Difference (TD) learning methods  \cite{sutton2018reinforcement}, \textit{SARSA} and \textit{Q-Learning}, were adapted to generate an optimal leader control policy. These methods' use of bootstrapping provides the flexibility needed to accommodate the stochastic nature of the follower agents' transitions between vertices. Additionally, TD methods are model-free approaches, which are suitable for our control objective since the leader does not have a model of the followers' behavior. We compare the two methods to identify their advantages and disadvantages when applied to our swarm herding problem. Our approach is based on the mean-field model \eqref{eq:ctrsys} in the sense that the leader learns a control policy using its observations of the population fractions of followers at all vertices in the graph.



Sutton and Barto \cite{sutton2018reinforcement} provide a formulation of the two TD algorithms that we utilize.
Let $S$ denote the state of the environment, defined later in this section; $A$ denote the action set of the leader, defined as the set $A_v$ in Equation \eqref{eq:action_set}; and $Q(S,A)$ denote the state-action value function. We define $\alpha \in [0, 1]$ and $\gamma \in [0, 1]$ as the learning rate and the discount factor, respectively. The policy used by the leader is determined by a state-action pair $(S,A)$.  $R$ denotes the  reward for the implemented policy's transition from the current to the next state-action pair and is defined in Equation \eqref{eq:reward}. In the SARSA algorithm, an on-policy method, the state-action value function is defined as: 
\begin{equation}
    \label{eq:SARSA_original}
    Q(S, A) \leftarrow Q(S, A) + \alpha[R + \gamma Q(S', A') - Q(S, A)]
\end{equation}
where the update is dependent on the current state-action pair $(S,A)$ and the next state-action pair $(S',A')$ determined by enacting the policy. In the Q-Learning algorithm, an off-policy method, the state-action value function is: 
\begin{equation}
    \label{eq:qlearning_original}
    Q(S, A) \leftarrow Q(S, A) + \alpha[R + \gamma \max_{a}Q(S', a) - Q(S, A)]
\end{equation} 
Whereas the SARSA algorithm update \eqref{eq:SARSA_original} requires knowing the next action $A'$ taken by the policy, the Q-learning update \eqref{eq:qlearning_original} does not require this information. 

Both algorithms use a discretization of 
the observed state $S$ and represent the  state-action value function $Q$ in tabular form 
as a multi-dimensional matrix, indexed by the leader actions and  states.
The state $S$ is 
defined as a vector that contains a discretized form of the population fraction of follower agents at each location $v \in \mathcal{V}$ and the location $\ell_1(k) \in \mathcal{V}$ of the leader agent. The leader's spatial state $\ell_1(k)$ must be taken into account because the leader's possible actions depend on its current location on the graph. Since the population fractions of follower agents are continuous values, we convert them into discrete integer quantities serving as a discrete function approximation of the continuous fraction populations. Instead of defining $F_v$ as the integer count of followers at location $v$, which could be very large, we reduce the dimensionality of the state space by discretizing the follower population fractions into $D$ intervals and scaling them up to integers between 1 and $D$:
\begin{equation}
    \label{eq:disretization}
    \begin{array}{l}
        F_{v} = \text{round} \left(\frac{D}{N}\sum_{i=1}^N\chi_v(X_i(0))\right), \\
        \text{where}~~ F_v \in [1, \ldots, D], ~~v \in \mathcal{V}.
    \end{array}
\end{equation}
For example, suppose $D = 10$. Then a follower  population fraction of 0.24 at location $v$ would have a corresponding state value $S_v = 2$.
Using a larger value of $D$ provides a finer classification of agent populations,  but at the cost of increasing the size of the state $S$. Given these definitions, the state vector $S$ is defined as:
\begin{equation}
    S_{env} = [F_{1},\ldots,F_{M}, \ell_1]
\end{equation}

The state vector $S_{env}$  contains many states that are inapplicable to the learning process. For example, the state vector for a $2 \times 2$ grid graph with $D = 10$ has $10 \times 10 \times 10 \times 10 \times 4$ possible variations, but only $10 \times 10 \times 10 \times 4$ are applicable since they satisfy the constraint that the follower population fractions at all vertices must sum up to $1$ (note that the sum $\sum_v F_v$ may differ slightly from $1$ due to the rounding used in Equation \eqref{eq:disretization}.) The new state $S_{env}$ is used as the state $S$ in the state-action value functions \eqref{eq:SARSA_original} and \eqref{eq:qlearning_original}.



The leader's control policy for both 
functions \eqref{eq:SARSA_original} and \eqref{eq:qlearning_original} is the following $\epsilon$-greedy policy, where $X \in [0, 1]$ is a uniform random variable 
and $\epsilon$ is a threshold parameter that determines the degree of state exploration during training:
\begin{equation}
    \label{eq:e_greedy_policy}
    \pi(S_{env}) = \argmax_{A}Q(S_{env}, A) \quad \text{if } X > \epsilon
\end{equation}


\section{Simulation Results}
\label{sec:simulations}

\begin{figure*}
	\centering
	\includegraphics[width=\textwidth]{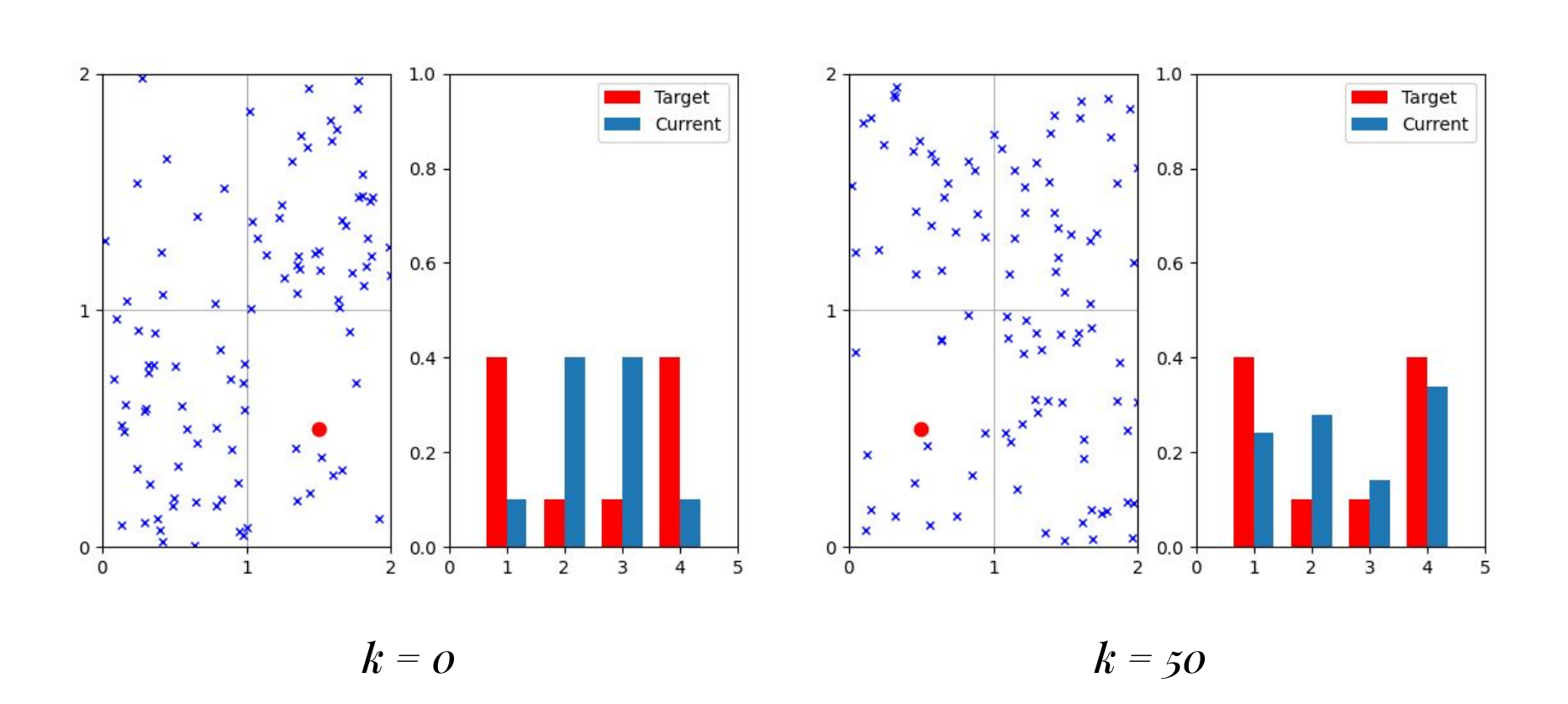} 
	\caption{Visual rendering of a  simulated scenario in our \textit{OpenAI} environment for iterations $k =$ 0 and 50.
	The environment simulates a strongly connected $2 \times 2$ grid graph such as the one shown in Figure  \ref{fig:markov_chain}. The leader (red circle) 
	moves between grid cells in a horizontal or vertical direction. It may not move diagonally. Follower agents (blue $\times$ symbols) 
	are randomly distributed in each cell. The borders of each cell are represented by the grid lines. The histogram to the right of each grid shows both the target (red) and current (blue) agent population fractions in each vertex at iteration $k$.}
	\label{fig:sim_example}
\end{figure*}

An \textit{OpenAI Gym} environment \cite{OpenAIGym16} was created in order to design, simulate, and visualize our leader-based  herding control policies \cite{KakishGym19}.
This open source virtual environment can be easily modified to simulate swarm controllers for different numbers of agents and graph vertices, making it a suitable environment for training  leader control policies using our model-free approaches.
The simulated controllers can then be implemented in physical robot experiments. Figure \ref{fig:sim_example} shows the simulated environment for a scenario with $100$ follower agents, represented by the blue $\times$ symbols, that are herded by a leader, shown as a red circle, 
over a $2 \times 2$ grid. The \textit{OpenAI} environment does not store the individual positions of each follower agent within a grid cell; instead, each cell is associated with an agent count. The renderer disperses agents randomly within a cell based on the cell's current agent count. 
The agent count for a grid cell is updated whenever an agent enters or leaves the cell according to the DTMC \eqref{eq:DTMC}, and the environment is re-rendered. Recording the agent counts in each cell rather than their individual positions significantly reduces 
memory allocation and 
computational time when training the leader control policy on scenarios with large numbers of agents. 

\begin{figure}[!t]
	\centering
    \includegraphics[height=45mm]{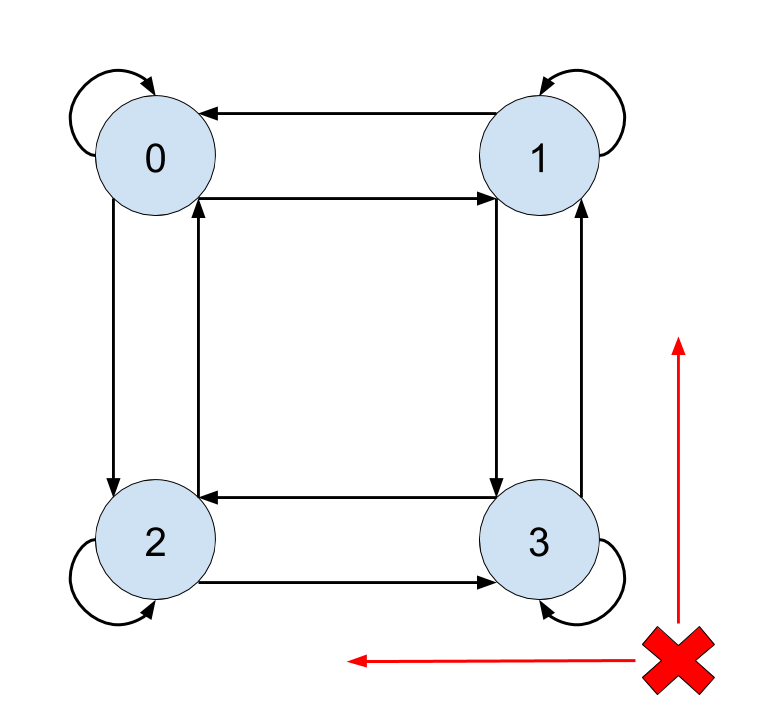}
	\caption{The bidirected grid graph $\mathcal{G}$ used in our simulated scenario. The leader agent (red $\times$ symbol) is 
	at vertex $3$. The movement options for the leader are \textit{Left} to vertex $2$ or \textit{Up} to vertex $1$. The leader can also \textit{Stay} at vertex $3$, where its presence triggers follower agents at the vertex to probabilistically transition to 
	vertex $1$ or vertex $2$.} 
	\label{fig:markov_chain}
\end{figure}

The graph $\mathcal{G}$ that models the 
environment in Figure \ref{fig:sim_example}, with each vertex of $\mathcal{G}$ corresponding to a grid cell, was defined as the $2 \times 2$ 
graph in Figure \ref{fig:markov_chain}. In the graph, agents transition along edges in either a horizontal or vertical direction, or they can stay at the current vertex.
The action set 
is thus defined as:
\begin{equation}
    \label{eq:grid_graph_action_set}
    A = [~\textit{Left}, ~\textit{Right}, ~\textit{Up}, ~\textit{Down}, ~\textit{Stay}~]
\end{equation}
 
Using the graph in Figure \ref{fig:markov_chain}, we trained and tested a leader control policy for follower agent populations of $N = 10$--$100$ at $10$-agent increments. Both the SARSA and Q-Learning paradigms were applied and trained on $5000$ episodes with $5000$ iterations each. In every episode, the initial distribution $\hat{S}_{initial}$ and target distribution $\hat{S}_{target}$ of the follower agents 
were defined as:
\begin{equation}
    \hat{S}_{initial} =
    \begin{bmatrix}
        0.4~&~0.1~&~0.1~&~0.4
    \end{bmatrix}
    ^{T}
    \label{eq:initialDist}
\end{equation}
\begin{equation}
    \hat{S}_{target} = 
    \begin{bmatrix}
        0.1~&~0.4~&~0.4~&~0.1
    \end{bmatrix}
    ^{T}
    \label{eq:targetDist}
\end{equation}
The initial leader location, $\ell_{1}$, was randomized to allow  many possible permutations of states $S_{env}$ for training. During training, an episode 
completes once the distribution of $N$ follower agents reaches a specified terminal state. Instead of defining the terminal state as the exact target distribution $\hat{S}_{target}$, which becomes more difficult to reach as $N$ increases due to the stochastic nature of the followers' transitions, we define this state as a distribution that is sufficiently close to $\hat{S}_{target}$. 
The learning rate and discount factor were set to $\alpha = 0.3$ and $\gamma = 0.9$, respectively. The follower agent transition rate $\beta_e$ was defined as the same value $\beta$ for all edges $e$ in the graph and
was set to 
$\beta=0.025$, $0.05$, or $0.1$. We use the mean squared error (MSE) to measure the difference between the current follower distribution and $\hat{S}_{target}$. 
The terminal state is reached when the MSE decreases below a 
 threshold value $\mu$. We trained our algorithms on 
threshold values of $\mu = 0.0005$, $0.001$, $0.0025$, and $0.005$. 

\begin{figure*}
    \centering
	\includegraphics[width=\textwidth]{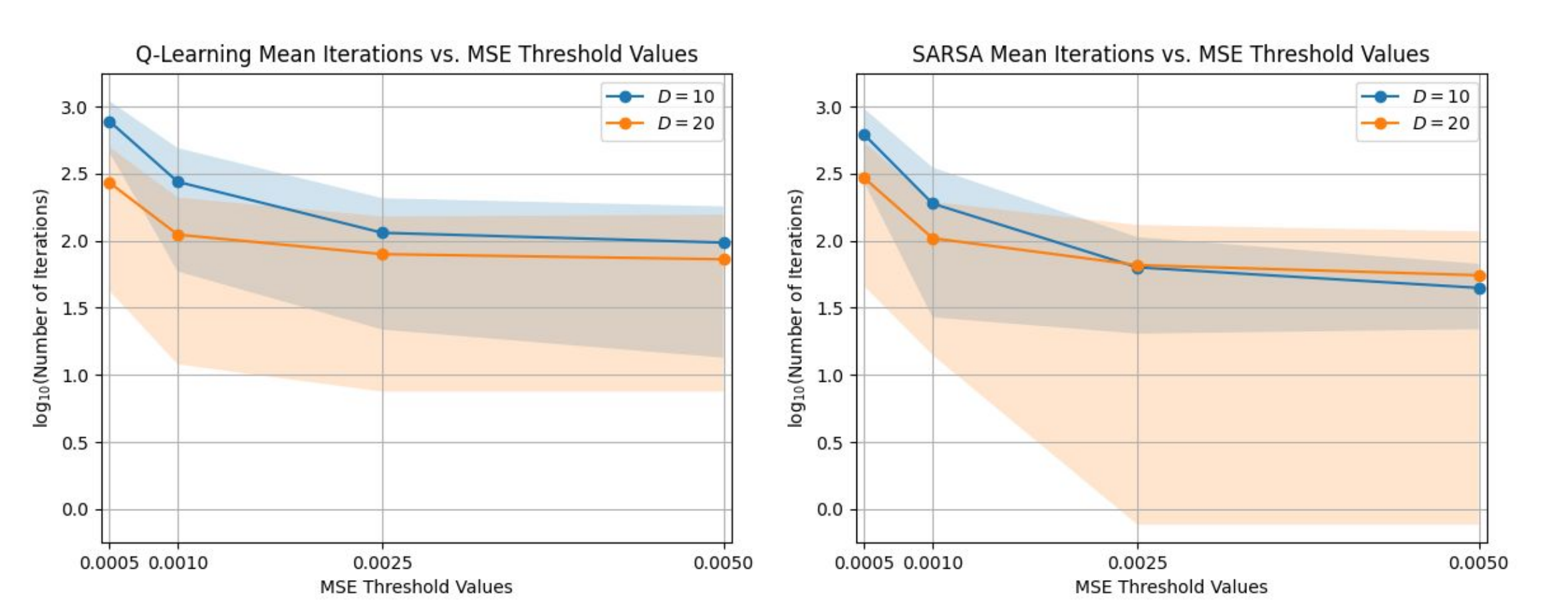}
	\caption{Number of iterations until convergence to $\hat{S}_{target}$ (plotted on a log scale) versus the MSE threshold value $\mu$ 
	for leader control policies that were learned using Q-Learning ({\it left}) and SARSA ({\it right}) with $\beta = 0.05$ and $N=100$ follower agents. 
    Each circle on the plots marks the mean number of iterations until convergence over 1000 test runs of a leader policy in the simulated grid graph environment in Figure \ref{fig:markov_chain}. The shaded regions indicate the range of $\pm$1 standard deviation about the mean numbers of iterations (blue for $D = 10$; orange for $D = 20$.) 
	}
	\label{fig:sarsa_q_learning_mse}
\end{figure*}

\begin{figure*}
    \centering
	\includegraphics[width=\textwidth]{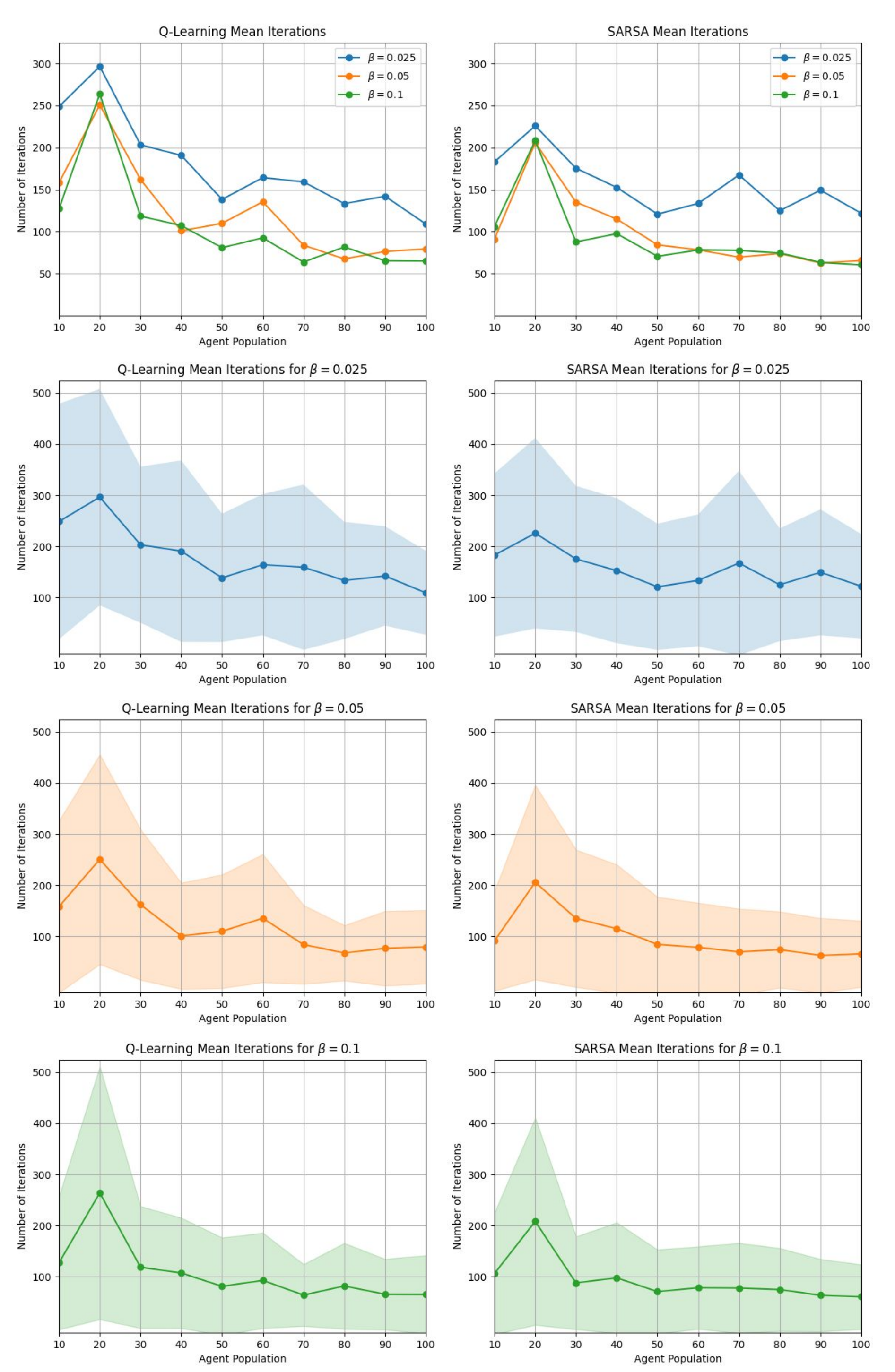}
 	\caption{Number of iterations until convergence to $\hat{S}_{target}$ versus number of follower agents $N$ for leader control policies that were learned using Q-Learning ({\it left column}) and SARSA ({\it right column}) with $D = 20$ and $\mu = 0.0025$. Each circle on the plots marks the mean number of iterations until convergence over 1000 test runs of a leader policy in the simulated grid graph environment in Figure \ref{fig:markov_chain} with the same value of $N$ that the policy was trained on.  The plot for each $\beta$ value in the top two figures are reproduced individually in the three figures below them, 
 	 along with shaded regions that indicate the range of $\pm$1 standard deviation about the mean numbers of iterations.  
 	}
 	\label{fig:sarsa_q_learning_results}
\end{figure*}

After training the leader control policies on each follower agent 
population size $N$, the policies were tested on scenarios with the same environment and value of $N$. The policy for each scenario was run $1000$ times to evaluate its performance. The policies were compared for terminal states that corresponded to the four different MSE threshold values $\mu$,
and were given $1000$ iterations to converge within the prescribed MSE threshold of the target distribution \eqref{eq:targetDist} from the initial distribution \eqref{eq:initialDist}.

Figure \ref{fig:sarsa_q_learning_mse} compares the performance of leader control policies that were designed using each TD algorithm as a function of the tested values of $\mu$. The leader control policies were trained on $N=100$ follower agents, using the parameters $\beta = 0.05$ and $D = 10$ or $20$, and tested in simulations with $N=100$.  
The plots show that for both policies, the mean number of iterations required to converge to $\hat{S}_{target}$ 
decreases as the threshold $\mu$ increases for constant $D$, and at low values of $\mu$, the mean number of iterations decreases when $D$ is increased.
In addition, as $\mu$ increases, the variance in the number of iterations
decreases (note the log scale of the $y$-axis in the plots)
or remains approximately constant, except for the $D=20$ case of SARSA. 


Figure \ref{fig:sarsa_q_learning_results} 
compares the performance of leader control policies that were designed using each algorithm as a function of 
$N$, where the leader policies were tested in simulations with the same value of $N$ that they were trained on. The other  parameters used for training were 
$\mu = 0.0025$, $D=20$, and $\beta = 0.025,$ $0.05$, or $0.1$.
 The figures show that raising $\beta$ from $0.05$ to $0.1$ does not significantly affect the 
 mean number of iterations until convergence, while decreasing $\beta$ from $0.05$ to $0.025$ results in a higher mean number of iterations. 
 This effect is evident for  both Q-Learning and SARSA trained leader control policies for $N > 50$. Both 
 leader control policies result in similar numbers of iterations for convergence at each agent population size. 
 Therefore, both the Q-Learning and SARSA training algorithms yield comparable performance 
for these scenarios.

The results in Figure  \ref{fig:sarsa_q_learning_results} show that as $N$ increases above $50$ agents, the mean number of iterations until convergence decreases slightly or remains approximately constant for all $\beta$ values and for $\mu=0.0025$. 
Moreover, from Figure \ref{fig:sarsa_q_learning_mse}, 
we see that MSE threshold values $\mu < 0.0025$ for $N=100$ result in a higher 
number of iterations 
than the $N=100$  case 
in Figure \ref{fig:sarsa_q_learning_results}. 
This trend may be 
due to differences in the magnitude of the smallest possible change in MSE over an iteration $k$ relative to the MSE threshold $\mu$ for different values of 
$N$. 
For example, for $N=10$, a similarity in iteration counts for all four MSE thresholds $\mu$ can be attributed to the fact that the change in the MSE due to a transition of one agent, corresponding to a change in population fraction of $1/N = 1/10$, 
is much higher than the four 
MSE thresholds (i.e., $(1/10)^{2} > 0.005$, $0.0025$, $0.001$, and $0.0005$). Compare this to the iteration count for $N=50$, which would have a corresponding change in MSE of 
$(1/50)^2$; this quantity is much smaller than $0.005$ and $0.0025$, 
but not much smaller than $0.001$ and $0.0005$. The iteration counts for $N=100$ are much lower, since $(1/100)^2$ is much smaller than all four MSE thresholds.

\begin{figure*}
    \centering
    \includegraphics[width=\textwidth]{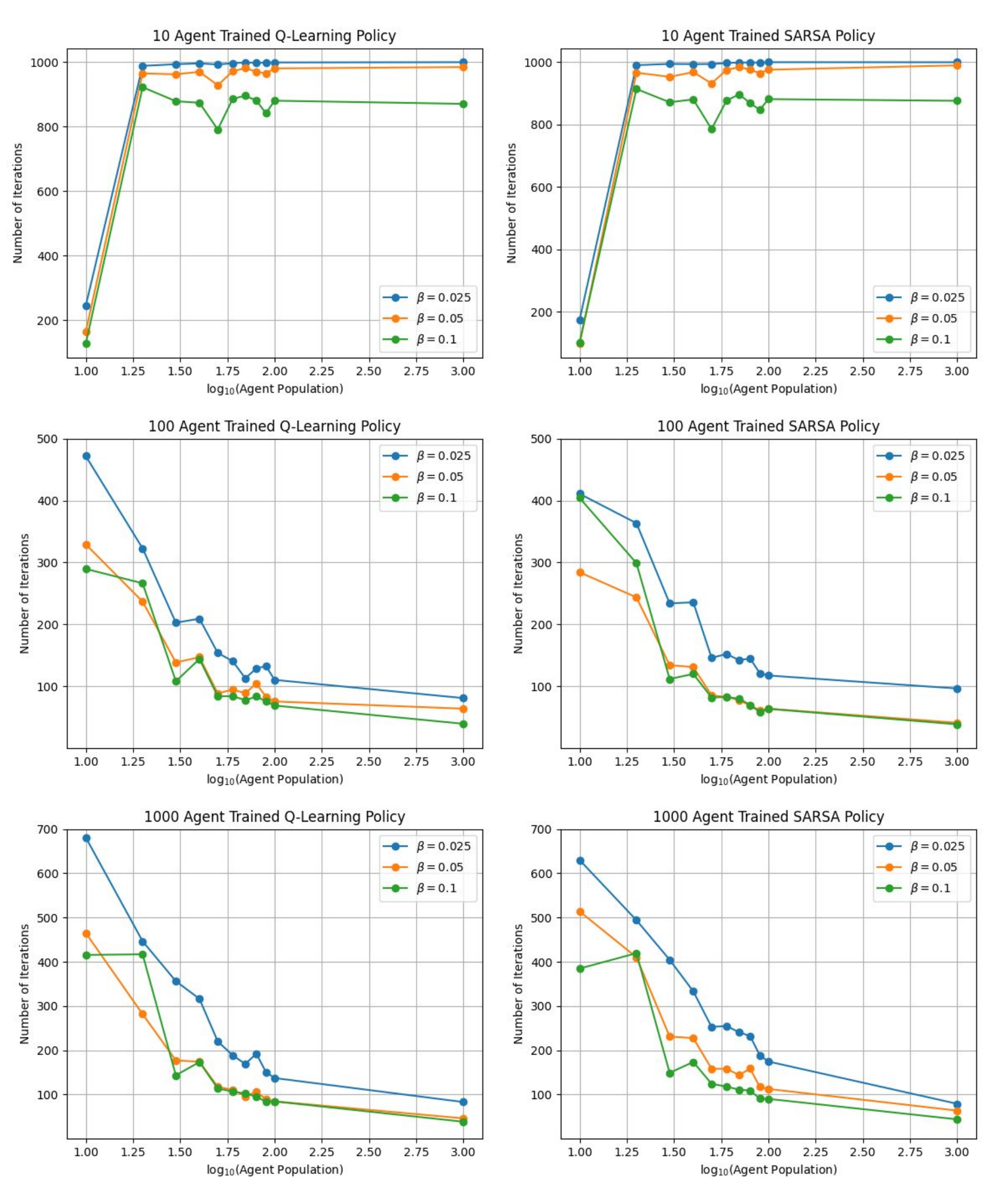}
    \caption{Number of iterations until convergence to $\hat{S}_{target}$ versus number of follower agents $N$ (plotted on a log scale) for leader control polices that were trained using Q-Learning ({\it left column}) and SARSA ({\it right column}) with $D = 20$; $\mu = 0.0025$; and $\beta = 0.025$, $0.05$, or $0.1$; and $N = 10$, $100$, or $1000$ agents. Each circle on the plots marks the mean number of iterations until convergence over 1000 test runs of a leader policy in the simulated grid graph environment in Figure \ref{fig:markov_chain}.
    }
    \label{fig:sim_results_with_10_100_1000}
\end{figure*}

Finally, Figure \ref{fig:sim_results_with_10_100_1000} compares the performance of leader control policies that were designed using each algorithm as a function of $N$, where the leader policies were trained with $N = 10$, $100$, or $1000$ follower agents and tested in simulations with $N = 10$--$100$ (at $10$-agent increments) and  $N=1000$ agents. This was done to evaluate the robustness of the policies trained on the three agents populations to changes in $N$. The other parameters used for training were
$\mu = 0.0025$, $D=20$, and $\beta = 0.025,$ $0.05$, or $0.1$. 
As the plots in Figure \ref{fig:sim_results_with_10_100_1000} show, policies trained on the smallest population, $N=10$, yield an increased mean number of iterations until convergence when applied to populations $N>10$. The reverse effect is observed, in general, for policies that are trained on higher values of $N$ than they are tested on. An exception is the case where the policies are trained on $N=100$ and $1000$ and tested on $N=10$, which produce much higher numbers of iterations than the policies that are both trained and tested on $N=10$. This is likely a result of the large variance, and hence greater uncertainty, in the time evolution of such a small agent population.
The lower amount of uncertainty in the time evolution of large swarms may make it easier for leader policies that are trained on large values of $N$ to control the distribution of a given follower agent population than policies that are trained on smaller values of $N$.
We thus hypothesize that training a leader agent with the mean-field model \eqref{eq:ctrsys} instead of the DTMC model would lead to improved performance in terms of a lower training time, since the policy would only need to be trained on one value of $N$, and fewer iterations until convergence to the target distribution.

\section{Experimental Results}
\label{sec:exp_results}



We also conducted experiments to verify that our herding approach is effective in a real-world environment with physical constraints on robot dynamics and inter-robot spacing. 
Two of the leader control policies that were generated in the simulated environment were tested on a swarm of small differential-drive robots in the \textit{Robotarium} 
\cite{WilsonRobotarium2020}, a remotely accessible swarm robotics testbed that provides an experimental platform for users to validate swarm algorithms and controllers.
Experiments are set up in the Robotarium with MATLAB or Python scripts. The robots move to target locations on the testbed surface using a position controller and avoid collisions with one another through the use of barrier certificates \cite{Wang2017}, a modification to the robots' controllers that satisfy particular safety constraints.
To implement this collision-avoidance strategy, the robots' positions and orientations in a global coordinate frame are measured from images taken from multiple \textit{VICON} motion capture cameras.

A video recording of our experiments is shown in \cite{physicalExperiment2020}.
The environment was represented as a $2 \times 2$ grid, as in the simulations, and $N = $ 10 robots were used as follower agents.  The leader agent, shown as the blue circle, and the boundaries of the four grid cells  were projected onto the surface of the testbed using an overhead projector. As in the simulations, at each iteration $k$, the leader moves from one grid cell to another depending on the action prescribed by its control policy. 
Both the SARSA and Q-Learning leader control policies trained with $N = 100$ follower agents, $D = 10$, $\mu = 0.0025$, and a $\beta = 0.1$ were implemented, and
 \cite{physicalExperiment2020} shows the performance of both control policies. In the video, the leader is red if it is executing the {\it Stay} action and blue if it is executing any of the other actions in the set $A$ (i.e., a movement action).
The current iteration $k$ and leader action are displayed at the top of the video frames. Actions that display $\epsilon$ next to them signify a random action as specified in \eqref{eq:e_greedy_policy}. 
 Each control policy was able to achieve the exact target distribution \eqref{eq:targetDist}. The SARSA method took $59$ iterations to reach this distribution, while the Q-Learning method took $23$ iterations. 



\section{Conclusion and Future Work}
\label{sec:conclusion}

We have presented two Temporal-Difference learning approaches to designing a leader-follower control policy for herding a swarm of agents as quickly as possible to a target  distribution over a graph. We demonstrated the effectiveness of the leader control policy in simulations and physical robot experiments for a range of swarm sizes $N$, illustrating the scalability of the control policy with $N$, and investigated the effect of $N$ on the convergence time to the target distribution. However, these approaches do not scale well with the graph size due to the computational limitations of tabular TD approaches and, in particular, our discretization of the system state into population fraction intervals. 
Our implementation requires a matrix with $D^{M} \times M \times |A|$ 
state-action values to train the leader control policy.
For our $2 \times 2$ grid graph with $|A|=5$ possible leader actions and $D = 20$ intervals, this is about $20^4 \times 4 \times 5$ values. 

To address this issue, our future work focuses on designing 
leader control policies using 
the mean-field model rather than the DTMC model for training, as suggested at the end of Section \ref{sec:simulations}. 
In this approach, the leader policies would be trained on follower agent population fractions that are computed from solutions of the discrete-time mean-field model 
\eqref{eq:ctrsys}, rather than from discrete numbers of agents that transition between locations according to a DTMC.
The leader control policies can also be modified to use function approximators such as neural networks for our training algorithm, allowing for utilization of modern deep reinforcement learning techniques. Neural network function approximators provide a more practical approach than tabular methods to improve the scalability of the leader control policy with graph size, in addition to swarm size. 
In addition, the control policies could be implemented on a swarm robotic testbed in a  \textit{decentralized} manner, in which each follower robot avoids collisions with other robots based on its local sensor information.  

\vspace{5mm}

\noindent \textbf{Acknowledgment} 
Many thanks to Dr. Sean Wilson at the Georgia Tech Research Institute for running the robot experiments on the Robotarium. 

%
%

\bibliographystyle{unsrt}
\bibliography{icra_bib}

\begin{thebibliography}{10}

\bibitem{sutton2018reinforcement}
Richard~S Sutton and Andrew~G Barto.
\newblock {\em Reinforcement learning: An introduction}.
\newblock MIT Press, 2018.

\bibitem{ji2008containment}
Meng Ji, Giancarlo Ferrari-Trecate, Magnus Egerstedt, and Annalisa Buffa.
\newblock Containment control in mobile networks.
\newblock {\em IEEE Transactions on Automatic Control}, 53(8):1972--1975, 2008.

\bibitem{mesbahi2010graph}
Mehran Mesbahi and Magnus Egerstedt.
\newblock {\em Graph theoretic methods in multiagent networks}, volume~33.
\newblock Princeton University Press, 2010.

\bibitem{pierson2017controlling}
Alyssa Pierson and Mac Schwager.
\newblock Controlling noncooperative herds with robotic herders.
\newblock {\em IEEE Transactions on Robotics}, 34(2):517--525, 2017.

\bibitem{elamvazhuthi2016confinement}
Karthik Elamvazhuthi, Sean Wilson, and Spring Berman.
\newblock Confinement control of double integrators using partially periodic
  leader trajectories.
\newblock In {\em American Control Conference}, pages 5537--5544, 2016.

\bibitem{paranjape2018robotic}
Aditya~A Paranjape, Soon-Jo Chung, Kyunam Kim, and David~Hyunchul Shim.
\newblock Robotic herding of a flock of birds using an unmanned aerial vehicle.
\newblock {\em IEEE Transactions on Robotics}, 34(4):901--915, 2018.

\bibitem{go2016reinforcement}
Clark~Kendrick Go, Bryan Lao, Junichiro Yoshimoto, and Kazushi Ikeda.
\newblock A reinforcement learning approach to the shepherding task using
  {SARSA}.
\newblock In {\em International Joint Conference on Neural Networks}, pages
  3833--3836, 2016.

\bibitem{elamvazhuthi2019mean}
Karthik Elamvazhuthi and Spring Berman.
\newblock Mean-field models in swarm robotics: {A} survey.
\newblock {\em Bioinspiration \& Biomimetics}, 15(1):015001, 2019.

\bibitem{vsovsic2018reinforcement}
Adrian {\v{S}}o{\v{s}}i{\'c}, Abdelhak~M Zoubir, and Heinz Koeppl.
\newblock Reinforcement learning in a continuum of agents.
\newblock {\em Swarm Intelligence}, 12(1):23--51, 2018.

\bibitem{huttenrauch2019deep}
Maximilian H{\"u}ttenrauch, Sosic Adrian, and Gerhard Neumann.
\newblock Deep reinforcement learning for swarm systems.
\newblock {\em Journal of Machine Learning Research}, 20(54):1--31, 2019.

\bibitem{yang2018mean}
Yaodong Yang, Rui Luo, Minne Li, Ming Zhou, Weinan Zhang, and Jun Wang.
\newblock Mean field multi-agent reinforcement learning.
\newblock In {\em International Conference on Machine Learning}, pages
  5567--5576, 2018.

\bibitem{OpenAIGym16}
Greg Brockman, Vicki Cheung, Ludwig Pettersson, Jonas Schneider, John Schulman,
  Jie Tang, and Wojciech Zaremba.
\newblock {OpenAI Gym}.
\newblock {\em arXiv preprint arXiv:1606.01540}, 2016.

\bibitem{KakishGym19}
Zahi Kakish.
\newblock {Herding OpenAI Gym Environment}, 2019.
\newblock \url{https://github.com/acslaboratory/gym-herding}.

\bibitem{WilsonRobotarium2020}
S.~{Wilson}, P.~{Glotfelter}, L.~{Wang}, S.~{Mayya}, G.~{Notomista}, M.~{Mote},
  and M.~{Egerstedt}.
\newblock The {R}obotarium: {G}lobally impactful opportunities, challenges, and
  lessons learned in remote-access, distributed control of multirobot systems.
\newblock {\em IEEE Control Systems Magazine}, 40(1):26--44, 2020.

\bibitem{Wang2017}
L.~{Wang}, A.~D. {Ames}, and M.~{Egerstedt}.
\newblock Safety barrier certificates for collisions-free multirobot systems.
\newblock {\em IEEE Transactions on Robotics}, 33(3):661--674, 2017.

\bibitem{physicalExperiment2020}
Z.~{Kakish}, K.~{Elamvazhuthi}, and S.~{Berman}.
\newblock Using reinforcement learning to herd a robotic swarm to a target
  distribution, 2020.
\newblock Autonomous Collective Systems Laboratory YouTube channel,
  \url{https://youtu.be/py3Pe24YDjE}.

\end{thebibliography}

%







\end{document}